\documentclass[11pt]{article}

\usepackage[final]{acl}

\usepackage{times}
\usepackage{latexsym}

\usepackage[T1]{fontenc}
\usepackage[utf8]{inputenc}

\usepackage{microtype}

\usepackage{inconsolata}

\usepackage{graphicx}
\usepackage{hyperref}
\usepackage{url}
\usepackage{amsmath}
\usepackage{amsfonts}
\usepackage{subcaption}
\usepackage{bm}
\usepackage{booktabs}
\usepackage{caption}
\usepackage{multirow}
\usepackage{array}
\usepackage{enumitem}
\usepackage{algorithm}
\usepackage{algpseudocode}
\usepackage{cleveref}
\usepackage{stfloats}
\usepackage{xcolor}
\usepackage{tcolorbox}
\usepackage{colortbl}
\usepackage{textcomp}
\usepackage{tabularx}
\usepackage{longtable}
\usepackage{dashrule}
\usepackage{relsize}
\usepackage{upquote}

\definecolor{lavender}{RGB}{230,230,250}
\definecolor{brickred}{RGB}{150,0,24}

\algrenewcommand{\algorithmiccomment}[1]{\hfill\textcolor{blue!50!white}{\small $\triangleright$ #1}}
\algtext*{EndIf}
\algtext*{EndWhile}

\newcommand{\method}{\textsc{HiDiffTIR}}

\title{\method{}: Hierarchical Difficulty-Aware Policy Optimization for Multi-Turn Tool-Integrated Reasoning}

\author{
 Yucan Guo\textsuperscript{1,2,3}\thanks{Co-first authors.},
 Xiaohan Wang\textsuperscript{4}$^{*}$,
 Miao Su\textsuperscript{1,2,3},
 Saiping Guan\textsuperscript{1,2,3}\thanks{Corresponding authors.},\\
 \textbf{Zhongni Hou\textsuperscript{4}},
 \textbf{Jiajun Chai\textsuperscript{4}},
 \textbf{Wei Lin\textsuperscript{4}},
 \textbf{Guojun Yin\textsuperscript{4}},\\
 \textbf{Xiaolong Jin\textsuperscript{1,2,3}$^{\dagger}$},
 \textbf{Jiafeng Guo\textsuperscript{1,2,3}},
 \textbf{Xueqi Cheng\textsuperscript{1,2,3}},
\\
    \textsuperscript{1}State Key Laboratory of AI Safety \\
    \textsuperscript{2}Institute of Computing Technology, Chinese Academy of Sciences \\
    \textsuperscript{3}School of Computer Science and Technology, University of Chinese Academy of Sciences \\
    \textsuperscript{4}Meituan\\
    \small
    \texttt{\{guoyucan23z, guansaiping, jinxiaolong\}@ict.ac.cn
    }\\
}

\begin{document}
\maketitle
\begin{abstract}
Tool-Integrated Reasoning (TIR) is a fundamental capability for LLM agents to solve complex tasks by interacting with external tools iteratively. Reinforcement Learning (RL) has become the dominant paradigm for enabling this capability. However, existing approaches typically assign uniform trajectory-level advantages and treat all correct tool calls equally, ignoring the varying difficulty and learning value across trajectories and reasoning steps. This can lead to imprecise learning signals that do not adequately distinguish between trivial and challenging tool-use patterns.
To address this limitation, we propose \method{}, a \textbf{Hi}erarchical \textbf{Diff}iculty-aware policy optimization framework for multi-turn \textbf{TIR}. \method{} performs difficulty-aware credit assignment at both trajectory and turn levels, enabling the policy to focus on more informative trajectories and harder reasoning steps. Notably, this fine-grained optimization is achieved without additional supervision, relying solely on group-level statistics derived from standard RL rollouts.
Extensive experiments on three tool-using benchmarks demonstrate that \method{} consistently improves multi-turn TIR performance and tool invocation accuracy over strong RL baselines, highlighting the necessity of difficulty-aware credit assignment for effective policy optimization in tool-integrated LLM agents. \footnote{The code is publicly available at \url{https://github.com/YucanGuo/HiDiffTIR}.}

\end{abstract}

\section{Introduction}
\label{sec:introduction}
Large Language Models (LLMs) have demonstrated strong reasoning capabilities across a wide range of tasks~\cite{Achiam2023gpt, Guo2025deepseek, Yang2025qwen3}, leading to the emergence of LLM agents that can plan, act, and interact with external environments~\cite{Huang2024understanding, Wang2024agentsurvey, Luo2025agentsurvey}. A central capability of such agents is effective tool use, where they invoke external tools to solve diverse tasks~\cite{Masterman2024landscape, Xu2025llm}.
Tool-Integrated Reasoning (TIR) formalizes this process by enabling LLM agents to iteratively interact with external tools~\cite{Gou2024tora, Qian2025toolrl}, such as APIs~\cite{Patil2024gorilla, Qin2024toolllm}, search engines~\cite{Li2025search, Jin2025search}, and code interpreters~\cite{Gou2024tora, Li2025torl}.
This paradigm extends LLM agents beyond their static parametric knowledge and their known limitations in domains such as precise numerical computation~\cite{Frieder2023mathematical}, using specific tools to solve complex multi-turn tasks that are otherwise challenging for purely inner knowledge-based reasoning~\cite{Lin2025understanding, Qu2025tool}.

Reinforcement learning (RL) has emerged as the prevailing paradigm for training LLMs to perform TIR~\cite{Qian2025toolrl, Xue2025simpletir}. 
By employing RL algorithms like Group Relative Policy Optimization (GRPO)~\cite{Shao2024deepseekmath}, models learn to interleave reasoning, tool invocation, and answer generation in an end-to-end manner.
Most existing TIR methods adopt trajectory-level optimization, which has evolved through two primary stages. 
Early methods rely on outcome-based rewards~\cite{Li2025torl, Dong2025tool}, assessing only the final correctness of a reasoning chain, and thus suffer from reward sparsity. 
To provide denser feedback, recent works have shifted toward turn-level verification~\cite{Qian2025toolrl, Ye2025feedback}, where the correctness of each individual tool-call is evaluated and subsequently aggregated to form the trajectory-level reward. Despite this progress, these methods typically assign uniform advantages across all steps within a trajectory, implicitly treating each tool call as equally informative for learning. 
This coarse-grained credit assignment can obscure the contribution of individual reasoning steps, especially in multi-turn settings where errors accumulate and only a subset of decisions are critical to success.

\begin{figure}[t]
    \centering
    \includegraphics[width=\linewidth]{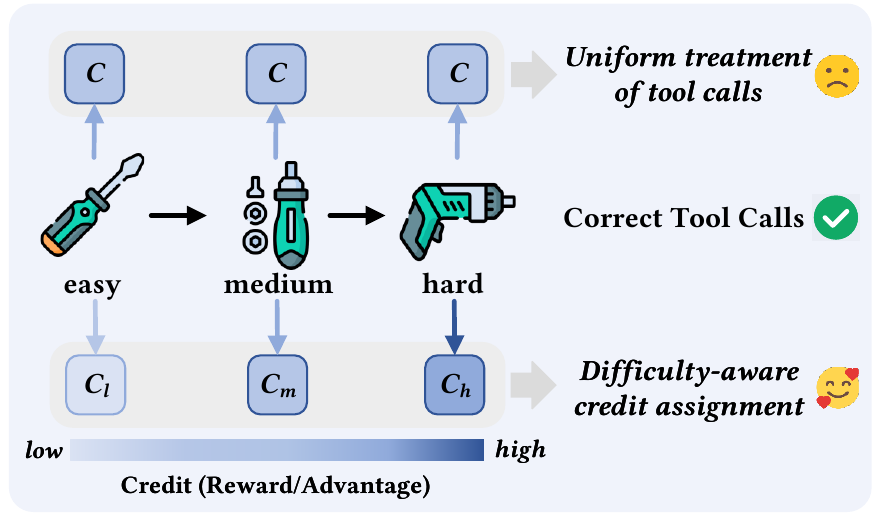}
    \caption{Comparison of credit assignment manners in TIR. Existing methods treat all correct tool calls uniformly, while \method{} performs difficulty-aware credit assignment.}
    \label{fig:motivation}
\end{figure}

To address this limitation, more recent works have explored finer-grained credit assignment by incorporating turn-level signals and combining them with trajectory-level objectives~\cite{Li2026deepagent, Qu2026matchtir}. These methods typically perform trajectory-turn fusion, where trajectory-level rewards are used in their original form, and turn-level signals are either directly applied or heuristically adjusted and propagated across steps.
However, they treat all correct tool calls equivalently, without explicitly modeling the varying difficulty of different reasoning steps. As a result, the learning signal remains insensitive to the relative importance and challenge of different tool-use patterns.

In this paper, we revisit credit assignment in TIR from a difficulty-aware perspective. As shown in~\Cref{fig:motivation}, we argue that not all trajectories or tool-calling steps contribute equally to policy improvement. Some trajectories are more informative due to their relative difficulty, and some reasoning steps are inherently harder and thus more valuable for learning. Motivated by this, we propose \method{}, a \textbf{Hi}erarchical \textbf{Diff}iculty-aware policy optimization framework for \textbf{TIR} that performs fine-grained credit assignment without requiring additional supervision.
\method{} operates at two levels. At the trajectory level, it differentiates rollouts based on their relative difficulty, enabling the policy to prioritize more informative training signals. At the turn level, it further refines credit assignment by redistributing advantages across reasoning steps according to their difficulty, allowing the model to focus on harder tool-calling decisions. Importantly, the judgment of tool-using difficulty relies solely on group-level statistics derived from standard RL rollouts and does not require external supervision such as LLM-based judges or rubrics. Our contributions are summarized as follows:
\begin{itemize}
    \item We propose \method{}, a hierarchical policy optimization framework that performs difficulty-aware credit assignment at both the trajectory and turn levels.
    \item We introduce a supervision-free method for fine-grained policy optimization that relies solely on statistics from standard RL rollouts.
    \item Extensive experiments demonstrate that \method{} improves tool-use accuracy and multi-turn reasoning performance across multiple benchmarks.
\end{itemize}

\section{Related Work}
\label{sec:related work}
\subsection{Tool-Integrated Reasoning}
TIR has become a central paradigm for equipping LLM agents with external tools to solve complex tasks. Early work primarily relies on Supervised Fine-Tuning (SFT), where models are trained on solution annotations from strong LLMs that interleave reasoning steps and tool invocations~\cite{Schick2023toolformer, Tang2023toolalpaca, Patil2024gorilla, Qin2024toolllm}. 
While effective in establishing basic tool-use capabilities, recent research has increasingly shifted toward RL to further improve policy learning through interaction~\cite{Xue2025simpletir, Li2025torl, Singh2025agentic}. 
Within RL-based TIR, methods have evolved from trajectory-level optimization with outcome-based rewards~\cite{Li2025torl, Dong2025tool} to incorporating process-based verification signals, such as tool name, parameter name, and argument values, to provide denser supervision~\cite{Qian2025toolrl, Ye2025feedback}. 
More recent approaches further integrate trajectory-level and turn-level signals to provide individual advantage for each turn. For instance, DeepAgent~\cite{Li2026deepagent} combines outcome rewards with turn-level tool call rewards, while MatchTIR~\cite{Qu2026matchtir} further propagates turn-level rewards across steps using discounted accumulation. 
Despite these advances, existing methods generally treat all correct tool calls as equally informative, without explicitly modeling the varying difficulty of different trajectories or reasoning steps.

\begin{figure*}
    \centering
    \includegraphics[width=\linewidth]{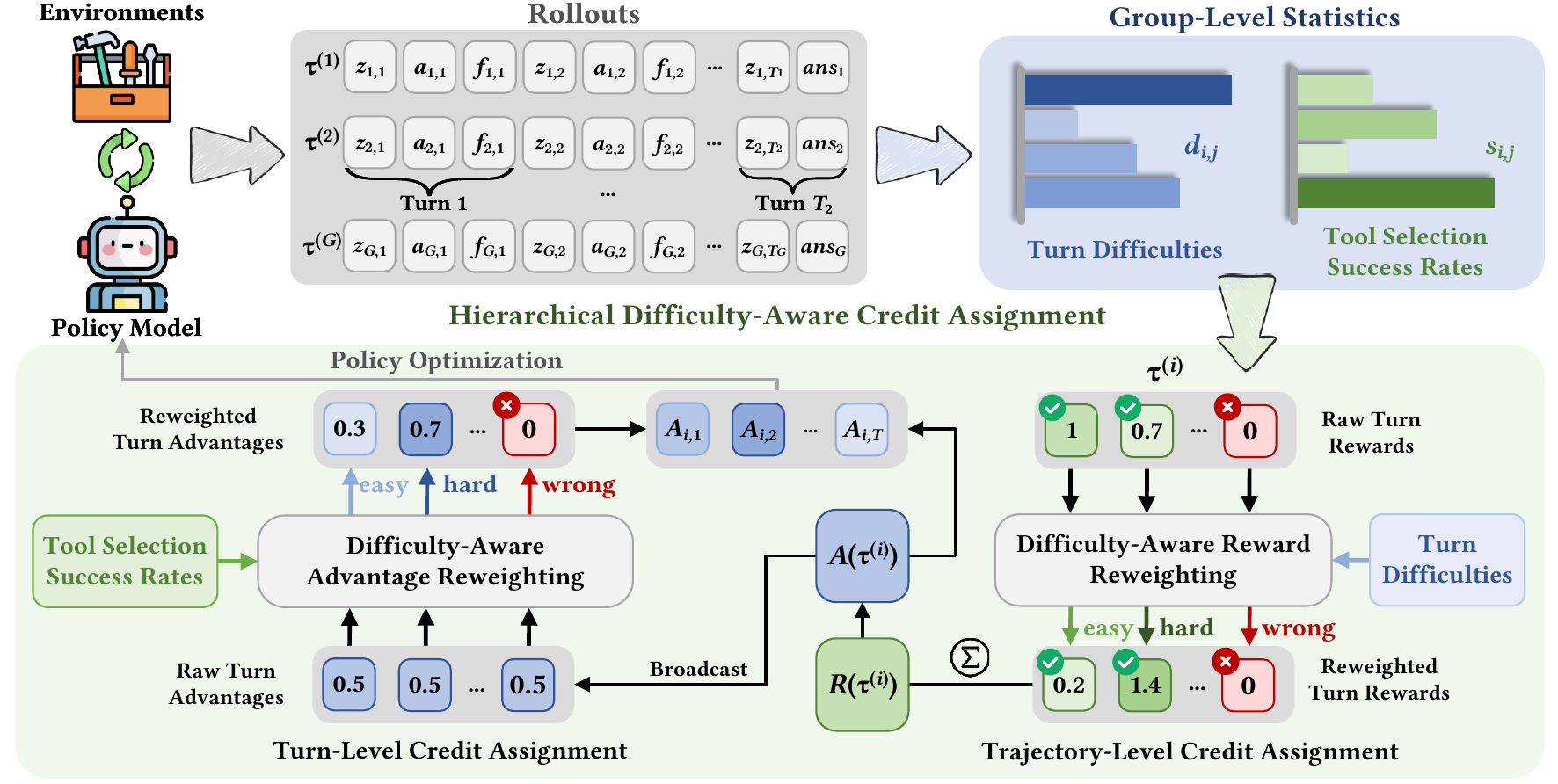}
    \caption{Illustration of the \method{} framework. \method{} implements a hierarchical difficulty-aware credit assignment mechanism: (a) trajectory-level, where raw turn rewards are reweighted based on turn difficulties to compute a difficulty-aware trajectory advantage, and (b) turn-level, where the trajectory advantage is redistributed to yield fine-grained reweighted turn advantages for policy optimization.}
    \label{fig:method}
\end{figure*}

\subsection{Reinforcement Learning for LLM Reasoning}
RL has been widely adopted to improve the reasoning capabilities of LLMs beyond SFT. A prominent line of work is Reinforcement Learning from Human Feedback (RLHF), where models are optimized using preference signals collected from human annotators~\cite{Christiano2017deep, Ouyang2022training}.
Recent work has demonstrated the effectiveness of Reinforcement Learning from Verifiable Rewards (RLVR), where models are optimized using automatically computed correctness signals~\cite{Shao2024deepseekmath, Guo2025deepseek, Yang2025qwen3}. 
To improve reward quality, a line of studies also concentrates on enhancing credit assignment in LLM reasoning through process-oriented supervision~\cite{Uesato2022solving, Lightman2023let}. Such approaches typically rely on additional supervision signals, including process reward models~\cite{Zhang2025lessons, Wang2025visualprm}, LLM-based judges guided by evaluation rubrics~\cite{Gunjal2025rubrics, Huang2025reinforcement}, to assess intermediate reasoning steps and provide more fine-grained feedback. 
In parallel, there is growing interest in difficulty-aware RL~\cite{Zhang2025grpolead, Gao2026divagrpo, Heckel2026asymmetric}, where training emphasizes more challenging samples to improve model performance, with sample difficulty estimated based on outcome-level signals.
However, these methods are largely designed for general reasoning tasks, making them insufficient for jointly modeling process-level feedback and difficulty in TIR.

\section{\method{}}
\subsection{Problem Formulation}
Given a user query $q$ and a set of available tools $\mathcal{T} = \{t_1, \dots, t_n\}$, the goal of a TIR agent $\pi_{\theta}$ is to solve the query through a multi-turn reasoning and tool use interaction process $\tau = \{(z_1, a_1, f_1), \dots, (z_{T-1}, a_{T-1}, f_{T-1}), (z_T, ans)\}$, where $T$ denotes the total number of turns.
At each intermediate turn $i < T$, the agent generates a reasoning step $z_i$ and a set of tool actions $a_i$, where each action specifies a selected tool from $\mathcal{T}$ along with its arguments. The environment executes these tool calls and returns feedback $f_i$. At the final turn, the agent produces a reasoning step $z_T$ followed by the final answer $ans$ without invoking any tool.
The objective of a TIR agent is to learn a policy that generates trajectories leading to correct final answers by effectively interleaving reasoning and tool use.

\subsection{Overall Framework}
\method{} is a hierarchical policy optimization framework for TIR that introduces difficulty-aware credit assignment at two levels. The core idea is to distinguish the relative learning importance of different trajectories and turns, rather than treating them uniformly.
As shown in~\Cref{fig:method}, at the trajectory level, \method{} distinguishes trajectories according to the relative difficulty of all constituent tool calls (\S\ref{sec:trajectory-level}). At the turn level, it further refines credit assignment across reasoning steps, considering both tool-selection difficulty and tool-using difficulty (\S\ref{sec:turn-level}). These two components together enable fine-grained optimization of the policy model (\S\ref{sec:policy optimization}).

\subsection{Trajectory-Level Difficulty-Aware Credit Assignment}
\label{sec:trajectory-level}
We now introduce trajectory-level difficulty-aware credit assignment, which extends turn-level rewards based on tool-use correctness. 
Specifically, we first compute the similarity between predicted and ground-truth tool calls to derive turn-level raw rewards. Next, we estimate the relative difficulty of each turn and reweight the raw rewards accordingly. Finally, we normalize these rewards to construct a difficulty-aware trajectory-level advantage.
 
\paragraph{Raw Reward Computation.} 
Following previous works~\cite{Qian2025toolrl, Qu2026matchtir}, we compute the raw reward for each turn in a trajectory using tool-level correctness. Given a trajectory $\tau^{(i)}$ with $T$ turns, let $C_i = \{c_{i1}, \dots, c_{im}\}$ denote the predicted tool calls and $C^*_i = \{c^*_{i1}, \dots, c^*_{in}\}$ the ground-truth calls. A similarity matrix $M_i \in \mathbb{R}^{m \times n}$ is defined as
\begin{equation}
    M_i[u,v] = \text{sim}(c_{iu}, c^*_{iv}),
\end{equation}
where $u\in\{1, \cdots, m\}$, $v\in\{1, \cdots, n\}$, and the similarity function $\text{sim}(\cdot, \cdot) \in [0,1]$ measures agreement on tool name, parameter names, and parameter values. Detailed computation of the similarity function is provided in Appendix~\ref{app:similarity function}.

A maximum-weight bipartite matching $\pi_i^*$ is then obtained via the Hungarian algorithm~\cite{Kuhn1955hungarian, Kuhn1956variants}:
\begin{equation}
    \pi_i^* = \arg\max_{\pi \in \Pi} \sum_{(u,v) \in \pi} M_i[u,v],
\end{equation}
where $\Pi$ denotes all one-to-one matchings between $C_i$ and $C^*_i$.

Let $C_{i,j} \subseteq C_i$ denote the subset of tool calls issued at turn $j$. The raw reward for turn $j$ is defined as
\begin{equation}
    r_{i,j} = \frac{1}{|C_{i,j}|} \sum_{c_{iu} \in C_{i,j}} \mathbb{I}[(u,v) \in \pi_i^*] \cdot M_i[u,v],
\end{equation}
where unmatched tool calls contribute zero.
These turn-level rewards serve as the initial supervision signal for subsequent trajectory-level optimization.

\paragraph{Difficulty-Aware Reward Reweighting.}
To model the relative difficulty of different tool-use behaviors, group-level statistics are estimated over trajectories sampled for the same query $q$. Let $\mathcal{G}(q) = \{\tau^{(1)}, \cdots, \tau^{(G)}\}$ denote the group of trajectories generated for $q$. For each ground-truth tool call $c^*$, we compute its average reward across the group:
\begin{equation}
    \bar{r}(c^*) = \frac{1}{|\mathcal{G}(q)|} \sum_{\tau^{(k)} \in \mathcal{G}(q)} r^{(k)}(c^*),
\end{equation}
where $r^{(k)}(c^*)$ denotes the reward assigned to the tool call in trajectory $\tau^{(k)}$ that is matched to $c^*$ via bipartite matching.

For each turn $j$ in trajectory $\tau^{(i)}$, let $C^*_{i,j}$ denote the set of ground-truth tool calls associated with that turn. We define the turn-level average reward as
\begin{equation}
    \bar{r}_{i,j} = \frac{1}{|C^*_{i,j}|} \sum_{c^* \in C^*_{i,j}} \bar{r}(c^*).
\end{equation}
The difficulty of the turn is then defined as
\begin{equation}
    d_{i,j} = 1 - \bar{r}_{i,j},
\end{equation}
and used to reweight the raw reward:
\begin{equation}
    \tilde{r}_{i,j} = d_{i,j} \cdot r_{i,j}.
\end{equation}
This weighting emphasizes turns associated with harder tool-use decisions while down-weighting consistently easy ones.

\paragraph{Trajectory-Level Score Construction.}
To ensure stable optimization, the reweighted rewards are normalized within each group. A min-max normalization is first applied:
\begin{equation}
    \hat{r}_{i,j} = \frac{\tilde{r}_{i,j} - \min_{\tau \in \mathcal{G}(q)} \tilde{r}_{\tau,j}}{\max_{\tau \in \mathcal{G}(q)} \tilde{r}_{\tau,j} - \min_{\tau \in \mathcal{G}(q)} \tilde{r}_{\tau,j}},
\end{equation}
followed by rescaling to preserve the total reward mass:
\begin{equation}
    \tilde{r}_{i,j}^{\text{norm}} = \hat{r}_{i,j} \cdot \frac{\sum_{\tau \in \mathcal{G}(q)} \tilde{r}_{\tau,j}}{\sum_{\tau \in \mathcal{G}(q)} \hat{r}_{\tau,j}}.
\end{equation}
The trajectory-level score is then obtained by aggregating normalized turn rewards:
\begin{equation}
    R(\tau^{(i)}) = \sum_{j=1}^{T_i} \tilde{r}_{i,j}^{\text{norm}},
\end{equation}
and converted into a group-relative advantage:
\begin{equation}
    A(\tau^{(i)}) = \frac{R(\tau^{(i)}) - \mu_q}{\sigma_q},
\end{equation}
where $\mu_q$ and $\sigma_q$ denote the mean and standard deviation of trajectory scores within $\mathcal{G}(q)$. This formulation yields a difficulty-aware trajectory-level objective that emphasizes relatively harder yet successful trajectories while maintaining stable comparisons within each group.

\subsection{Turn-Level Difficulty-Aware Credit Assignment}
\label{sec:turn-level}
While trajectory-level optimization captures the overall difficulty of a rollout, it does not distinguish which reasoning steps are more challenging within the trajectory. To further refine credit assignment, we first estimate the relative difficulty of each turn based on group-level statistics, and then redistribute the trajectory-level advantage across individual turns based on these turn-level difficulties.

\paragraph{Group-Level Tool Selection Success Rate.}
For each ground-truth tool call $c^*$, we compute its tool selection success rate over the group $\mathcal{G}(q)$:
\begin{equation}
    s(c^*) = \frac{1}{|\mathcal{G}(q)| + \lambda} \sum_{\tau^{(k)} \in \mathcal{G}(q)} \mathbb{I}\big(r^{(k)}(c^*) > 0\big) + \lambda,
\end{equation}
where $\mathbb{I}(\cdot)$ is an indicator function, and $\lambda$ is a smoothing constant. Since a positive reward indicates that the correct tool is selected, $s(c^*)$ effectively measures how frequently each tool call is correctly invoked across trajectories.

After that, we aggregate the tool selection success rates into a turn-level statistic for each turn $j$ in trajectory $\tau^{(i)}$:
\begin{equation}
    s_{i,j} = \min_{c^* \in C^*_{i,j}} s(c^*),
\end{equation}
which emphasizes the most challenging tool call within the turn.

\paragraph{Difficulty-Aware Advantage Reweighting.}
Based on the estimated tool selection success rates, we construct turn-level weights that emphasize more difficult decisions. For trajectories with positive advantage, we assign higher weights to turns with lower success rates while incorporating the overall tool-using quality:
\begin{equation}
    \tilde{w}_{i,j} = \frac{r_{i,j}}{\sqrt{s_{i,j}}}.
\end{equation}
This formulation suppresses incorrect turns while prioritizing harder correct decisions.

For trajectories with negative advantage, we do not apply difficulty-based reweighting and instead use uniform weights, ensuring stable penalization of unsuccessful rollouts. The weights are then normalized within each trajectory:
\begin{equation}
    \hat{w}_{i,j} = \frac{\tilde{w}_{i,j}}{\frac{1}{T_i} \sum_{k=1}^{T_i} \tilde{w}_{i,k}}.
\end{equation}

Finally, the trajectory-level advantage is redistributed across turns as
\begin{equation}
    A_{i,j} = A(\tau^{(i)}) \cdot \left( \alpha + (1 - \alpha)\hat{w}_{i,j} \right),
\end{equation}
where $\alpha \in [0,1]$ controls the balance between trajectory- and turn-level difficulty-aware credit assignment, preserving the global optimization signal while enabling fine-grained emphasis on harder reasoning turns.

\subsection{Policy Optimization}
\label{sec:policy optimization}
The final advantage assigned to each token is constructed by combining trajectory-level and turn-level signals in a hierarchical manner. 
The turn-level advantages $A_{i,j}$ are assigned to all tokens within the corresponding turn, resulting in token-level advantages $\tilde{A}_{i,t}$.
Using the integrated advantage $\tilde{A}_{i,t}$, we optimize the policy under the GRPO algorithm. Given a batch of queries $q \sim \mathcal{D}$ and sampled trajectories $\{\tau_i\}_{i=1}^G \sim \pi_{\theta_{\text{old}}}(\cdot \mid q)$, the objective is defined as
%
%\begin{small}
\begin{equation}
\begin{split}
\mathcal{J}(\theta) = \mathbb{E}_{q, \{\tau_i\}} \frac{1}{G} \sum_{i=1}^{G} \frac{1}{|\tau_i|} \sum_{t=1}^{|\tau_i|}
\bigg[
\min\Big(
w_{i,t} \tilde{A}_{i,t}, \\
\text{clip}(w_{i,t}, 1-\epsilon, 1+\epsilon) \tilde{A}_{i,t}
\Big)
- \beta \mathbb{D}_{\mathrm{KL}}(\pi_{\theta} \,\|\, \pi_{\text{ref}})
\bigg],
\end{split}
\end{equation}
%\end{small}
%
where $w_{i,t} = \frac{\pi_{\theta}(y_{i,t} \mid y_{i,<t}, q)}{\pi_{\theta_{\text{old}}}(y_{i,t} \mid y_{i,<t}, q)}$ denotes the token-level importance sampling ratio, and $\beta$ controls the strength of KL regularization.
This objective maintains the standard GRPO optimization structure while incorporating difficulty-aware hierarchical credit assignment, enabling more effective learning for multi-turn TIR.

\section{Experiments}
In this section, we evaluate \method{} through comparative experiments (\S\ref{sec:main results}), ablation study (\S\ref{sec:ablation study}), and tool-use accuracy analysis (\S\ref{sec:tool-use accuracy analysis}), with additional results detailed in Appendix~\ref{app:additional_experiments}.
\begin{table*}[ht]
	\centering
    \caption{Performance comparison between \method{} and the baselines on in-domain and out-of-domain benchmarks across Qwen3-4B and Qwen3-8B backbones. The best results within each backbone group are indicated in bold, while the underlined values represent the second-best results.}
	\label{tab:main results}
        \resizebox{1.0\textwidth}{!}{
	\begin{tabular}{l|cccc|cccccc}
        \toprule
        \multirow{3}{*}{\textbf{Methods}}
        & \multicolumn{4}{c|}{\textbf{In-Domain (ID)}}
        & \multicolumn{6}{c}{\textbf{Out-of-Domain (OOD)}} \\
        \cmidrule(lr){2-5} \cmidrule(lr){6-11}
        
        & \multicolumn{4}{c|}{\textbf{FTRL}}
        & \multicolumn{4}{c}{\textbf{BFCL}}
        & \textbf{ToolHop} 
        & \multirow{2}{*}{\textbf{Avg.}} \\
        \cmidrule(lr){2-5} \cmidrule(lr){6-9} \cmidrule(lr){10-10}
        
        & Solve-P & Solve-R & Solve-F1 & \textbf{Avg.}
        & Base & MF & MP & LC
        & AC & \\
        \midrule
        \rowcolor{gray!10}
        \multicolumn{11}{c}{{\centering\raisebox{-.3\height}{\includegraphics[width=0.4cm]{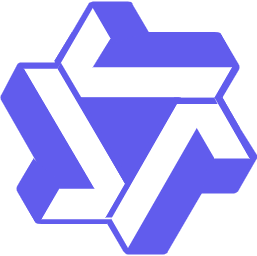}}} \textit{\textbf{Qwen3-4B}}}\\
        \midrule
        Vanilla
        & 30.78 & 29.65 & 25.85 & 28.76
        & 41.50 & 31.00 & 26.50 & 27.50
        & 31.63 & 31.63 \\
        
        GRPO
        & 31.13 & 32.83 & 30.67 & 31.54
        & 45.00 & 37.50 & 26.50 & 29.50
        & 37.25 & 35.15 \\
        
        ToolRL
        & 28.26 & 28.32 & 23.78 & 26.79
        & 32.50 & 31.00 & 22.50 & 20.00
        & 30.28 & 27.26 \\
        
        FTRL
        & \underline{34.10} & 34.07 & 31.54 & 33.24
        & 43.00 & 35.50 & \underline{31.00} & 28.00
        & 38.63 & 35.23 \\
        
        MatchTIR (OT)
        & 31.79 & 37.52 & 32.60 & 33.97
        & 50.00 & \underline{40.50} & 26.50 & 35.00
        & 41.95 & 38.79 \\
        
        MatchTIR (KM)
        & 32.39 & \underline{39.70} & \underline{34.21} & \underline{35.43}
        & \underline{50.50} & \textbf{47.00} & 28.50 & \underline{36.50}
        & \underline{42.55} & \underline{41.01} \\
        
        \midrule
        \rowcolor{lavender}
        \textbf{\method{}}
        & \textbf{35.41} & \textbf{44.74} & \textbf{38.29} & \textbf{39.48}
        & \textbf{51.00} & 39.50 & \textbf{31.50} & \textbf{37.00}
        & \textbf{49.95} & \textbf{41.79} \\
        
        \midrule
        \rowcolor{gray!10}
        \multicolumn{11}{c}{{\centering\raisebox{-.3\height}{\includegraphics[width=0.4cm]{figs/logo/qwen.png}}} \textit{\textbf{Qwen3-8B}}}\\
        \midrule
        
        Vanilla
        & 28.08 & 36.55 & 29.74 & 31.46
        & 47.50 & 46.00 & 37.50 & 34.50
        & 42.21 & 41.54 \\
        
        GRPO
        & 31.59 & 39.75 & 32.54 & 34.63
        & 52.50 & 45.50 & 34.50 & 36.50
        & 40.64 & 41.93 \\
        
        ToolRL
        & 25.57 & 35.31 & 26.72 & 29.20
        & 41.00 & 39.50 & 31.50 & 25.00
        & 32.93 & 33.99 \\
        
        FTRL
        & 32.32 & 38.87 & 32.85 & 34.68
        & 51.50 & 45.00 & 35.50 & 34.00
        & 36.72 & 40.54 \\
        
        MatchTIR (OT)
        & 33.61 & 42.56 & 33.61 & 36.59
        & 55.50 & \textbf{52.00} & 38.50 & 36.00
        & 45.80 & 45.56 \\
        
        MatchTIR (KM)
        & \underline{36.33} & \underline{44.18} & \underline{37.33} & \underline{39.28}
        & \underline{60.00} & \underline{49.00} & \underline{39.00} & \underline{40.50}
        & \underline{46.16} & \underline{46.93} \\

        \midrule
        \rowcolor{lavender}
        \textbf{\method{}}
        & \textbf{36.90} & \textbf{44.82} & \textbf{39.04} & \textbf{40.25}
        & \textbf{61.50} & \underline{49.00} & \textbf{40.00} & \textbf{42.00}
        & \textbf{51.26} & \textbf{48.75} \\
        \bottomrule
        \end{tabular}
	}
	
\end{table*}
\subsection{Experimental Setting}
\paragraph{Training Dataset.}
We conduct training on the FTRL dataset~\cite{Ye2025feedback}, which comprises 2,215 training data generated via an automated environment construction pipeline. By executing all tools locally as code, it circumvents the unreliability of online APIs, providing stable and verifiable tool-use training.

\paragraph{Evaluation Datasets.}
We evaluate \method{} on three widely used TIR benchmarks: 
(1) \textbf{FTRL}~\cite{Ye2025feedback}, the held-out test dataset generated by the FTRL pipeline;
(2) \textbf{ToolHop}~\cite{Ye2025toolhop}, a query-driven benchmark specifically designed to evaluate multi-hop tool reasoning; and 
(3) the \textbf{Berkeley Function Call Leaderboard (BFCL)}~\cite{Patil2025bfcl}, a comprehensive and executable function call benchmark that evaluates function-calling ability across diverse APIs and parameter configurations.

\paragraph{Baselines.}
We compare \method{} against several strong TIR baselines:
(1) \textbf{Vanilla}, which directly performs TIR using the backbone Qwen3~\cite{Yang2025qwen3} model without RL training;
(2) \textbf{GRPO}~\cite{Shao2024deepseekmath}, the standard GRPO framework, which computes advantages solely based on the outcome reward;
(3) \textbf{ToolRL}~\cite{Qian2025toolrl}, an RL-based TIR training framework that introduces process-based tool-call verification rewards;
(4) \textbf{FTRL}~\cite{Ye2025feedback}, a feedback-driven training framework using a verifiable reward mechanism based on precision and completeness;
and (5) \textbf{MatchTIR}~\cite{Qu2026matchtir}, a recent fine-grained credit assignment framework that derives turn-level rewards with two credit assignment manners: \textbf{MatchTIR (KM)}, which formulates the assignment as a bipartite matching problem using the Kuhn-Munkres algorithm~\cite{Kuhn1955hungarian}, and \textbf{MatchTIR (OT)}, which smooths the assignment through optimal transport~\cite{Cuturi2013sinkhorn}.

\paragraph{Implementation Details.}
We conduct experiments using Qwen3-4B and Qwen3-8B~\cite{Yang2025qwen3} as the backbone models. 
All RL training is implemented based on the verl framework~\cite{Sheng2025hybridflow}, with vLLM~\cite{Kwon2023efficient} accelerating the rollout generation.
For GRPO settings, we sample a group size of 16 responses per prompt at a temperature of 1.0. The policy model is trained for 5 epochs using a global batch size of 256, restricting the multi-turn reasoning process to a maximum of 10 turns per trajectory. 
%
% For our hierarchical advantage redistribution, 
We set the smoothing constant $\lambda$ to 1.0 and the fusion coefficient $\alpha$ to 0.7. All experiments are executed on a single node equipped with 8 NVIDIA H20 GPUs.
Additional implementation details are provided in Appendix~\ref{app:implementation details}.

\subsection{Main Results}
\label{sec:main results}
\begin{table*}[ht]
	\centering
    \caption{Ablation study on credit assignment using Qwen3-4B.}
	\label{tab:ablation}
        \resizebox{1.0\textwidth}{!}{
	\begin{tabular}{l|cccc|cccccc}
        \toprule
        \multirow{2}{*}{\textbf{Methods}}
        
        & \multicolumn{4}{c|}{\textbf{FTRL}}
        & \multicolumn{4}{c}{\textbf{BFCL}}
        & \textbf{ToolHop} 
        & \multirow{2}{*}{\textbf{Avg.}} \\
        \cmidrule(lr){2-5} \cmidrule(lr){6-9} \cmidrule(lr){10-10}
        
        & Solve-P & Solve-R & Solve-F1 & \textbf{Avg.}
        & Base & MF & MP & LC
        & AC & \\
        \midrule
        \rowcolor{lavender}
        \textbf{\method{}}
        & \textbf{35.41} & \textbf{44.74} & \textbf{38.29} & \textbf{39.48}
        & 51.00 & 39.50 & \textbf{31.50} & \textbf{37.00}
        & 49.95 & \textbf{41.79} \\
        \textit{w/o} trajectory-level & 33.65 & 40.69 & 35.77 & 36.70 & 49.50 & \textbf{40.00} & 30.00 & 32.50 & 51.36 & 40.67\\
        \textit{w/o} turn-level & 33.43 & 42.45 & 36.13 & 37.34 & \textbf{51.50} & 38.50 & 29.00 & 34.50 & \textbf{51.96} & 41.09\\

        \bottomrule
        \end{tabular}
	}
\end{table*}
The main experimental results are presented in~\Cref{tab:main results}. On the basis of these results, we make the following observations.

\textbf{\method{} consistently achieves the best overall performance across different evaluation benchmarks.}
\method{} outperforms all baselines on the average score for both Qwen3-4B and Qwen3-8B. Compared with existing RL-based TIR methods, our method demonstrates greater improvements in both in-domain and out-of-domain settings. In particular, our method consistently surpasses recent strong baselines such as FTRL and MatchTIR, showing that incorporating hierarchical difficulty-aware optimization provides complementary benefits beyond existing process-based and turn-level reward modeling strategies.

\textbf{The effectiveness of \method{} generalizes across different backbone sizes.}
Although larger backbones generally achieve stronger overall performance, \method{} consistently improves upon the corresponding baselines under both the 4B and 8B settings. Notably, the performance gains on the in-domain FTRL benchmark are more pronounced for Qwen3-4B, suggesting that smaller models benefit more from fine-grained difficulty-aware optimization. This observation indicates that improved credit assignment can partially compensate for weaker intrinsic reasoning and tool-use capabilities in smaller models. 
Meanwhile, the consistent improvements on Qwen3-8B demonstrate the robustness and scalability of our framework.

\textbf{\method{} brings larger improvements on datasets with a larger number of tools.}
As shown in~\Cref{tab:dataset_statistics}, datasets such as FTRL and ToolHop contain substantially larger numbers of tools compared with BFCL. Correspondingly, \method{} achieves more evident improvements on these benchmarks, which require reasoning over thousands of candidate tools. In contrast, on datasets with relatively fewer tools, such as BFCL, the performance improvements are comparatively modest, while stronger backbone models already obtain noticeable benefits from scaling alone. These findings highlight that the advantages of difficulty-aware optimization become increasingly important as the complexity of the tool environment grows.

\textbf{Current RL-based TIR methods still struggle with missing-information scenarios.}
Although most methods achieve clear improvements on the BFCL benchmark, the gains on the Missing Functions (MF) and Missing Parameters (MP) subsets remain modest, as they require the model to recognize when the currently available tools or user-provided information are insufficient. 
Existing RL-based TIR training mainly emphasizes successful tool execution and task completion, which may bias the policy toward aggressively invoking tools rather than learning insufficiency detection. 
Specifically, because our difficulty-aware reweighting amplifies successful complex tool sequences, it inadvertently reinforces this aggressive calling bias in sparse tool settings. Consequently, while \method{} significantly outperforms most baselines on the MF subset, it slightly trails MatchTIR. 
This observation suggests an important future direction for TIR, i.e., integrating self-awareness of capability boundaries into RL optimization. 

\subsection{Ablation Study}
\label{sec:ablation study}
We further conduct ablation studies on Qwen3-4B to evaluate the contribution of each component in \method{}. 
\paragraph{Effectiveness of Hierarchical Components.} As shown in~\Cref{tab:ablation}, removing either trajectory-level or turn-level difficulty-aware credit assignment leads to consistent performance degradation compared with the full model, demonstrating that both components contribute to the final performance. In particular, removing trajectory-level optimization results in a larger overall drop on FTRL, indicating that distinguishing the relative difficulty of different trajectories provides an important optimization signal for RL training. 
Furthermore, performance variations on Out-Of-Domain (OOD) datasets like BFCL and ToolHop are less pronounced when omitting specific components. Since our difficulty signals are derived dynamically during training, they naturally target bottlenecks within the training distribution. Thus, these components are highly effective for seen tools but have a more constrained impact on completely unseen tools in OOD scenarios, although the full \method{} model still achieves the best overall average performance.
\begin{figure}[t]
    \begin{subfigure}{0.23\textwidth}
        \includegraphics[width=\textwidth]{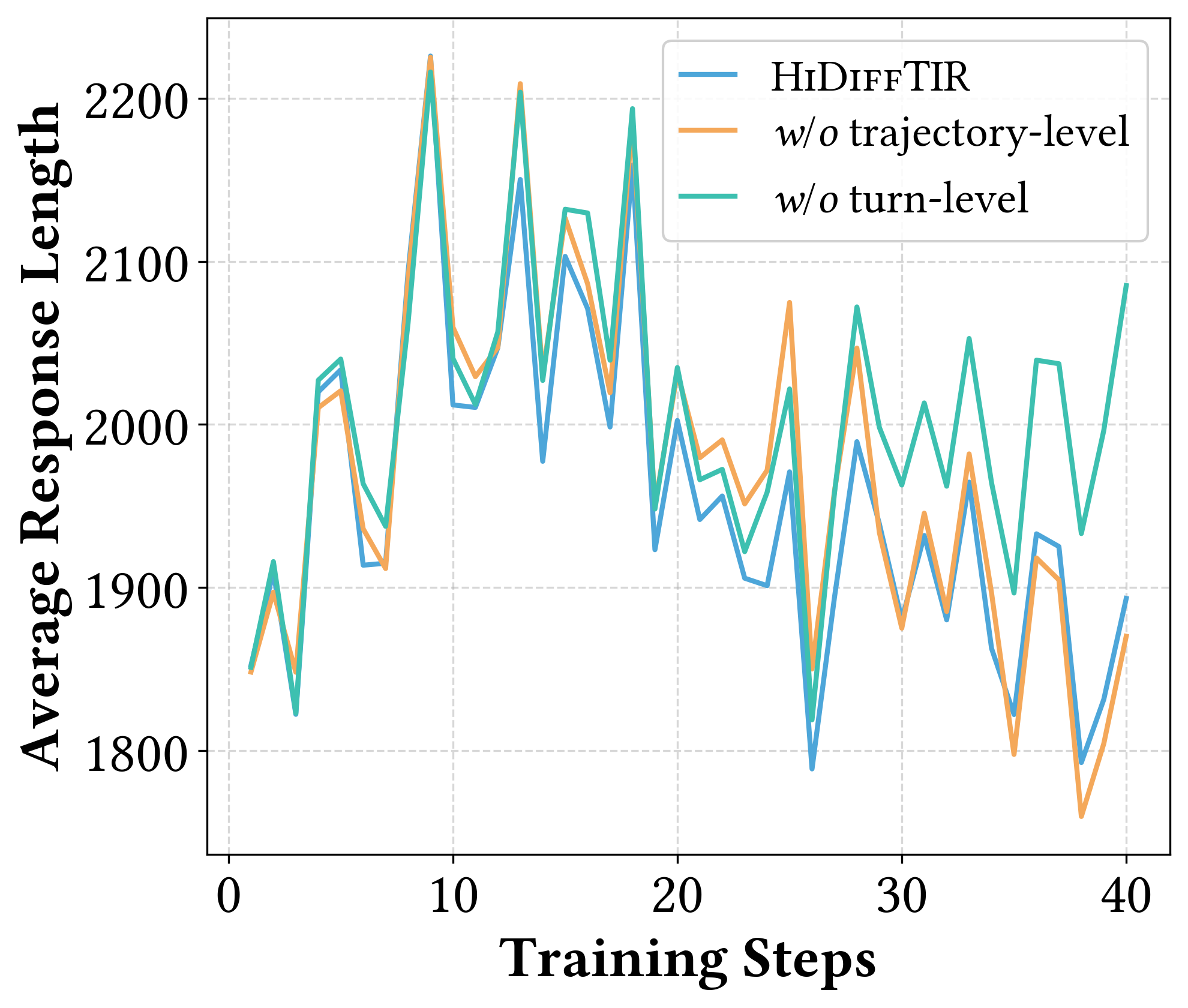}
        \caption{Average Response Length}
    \end{subfigure}
    \begin{subfigure}{0.23\textwidth}
        \includegraphics[width=\textwidth]{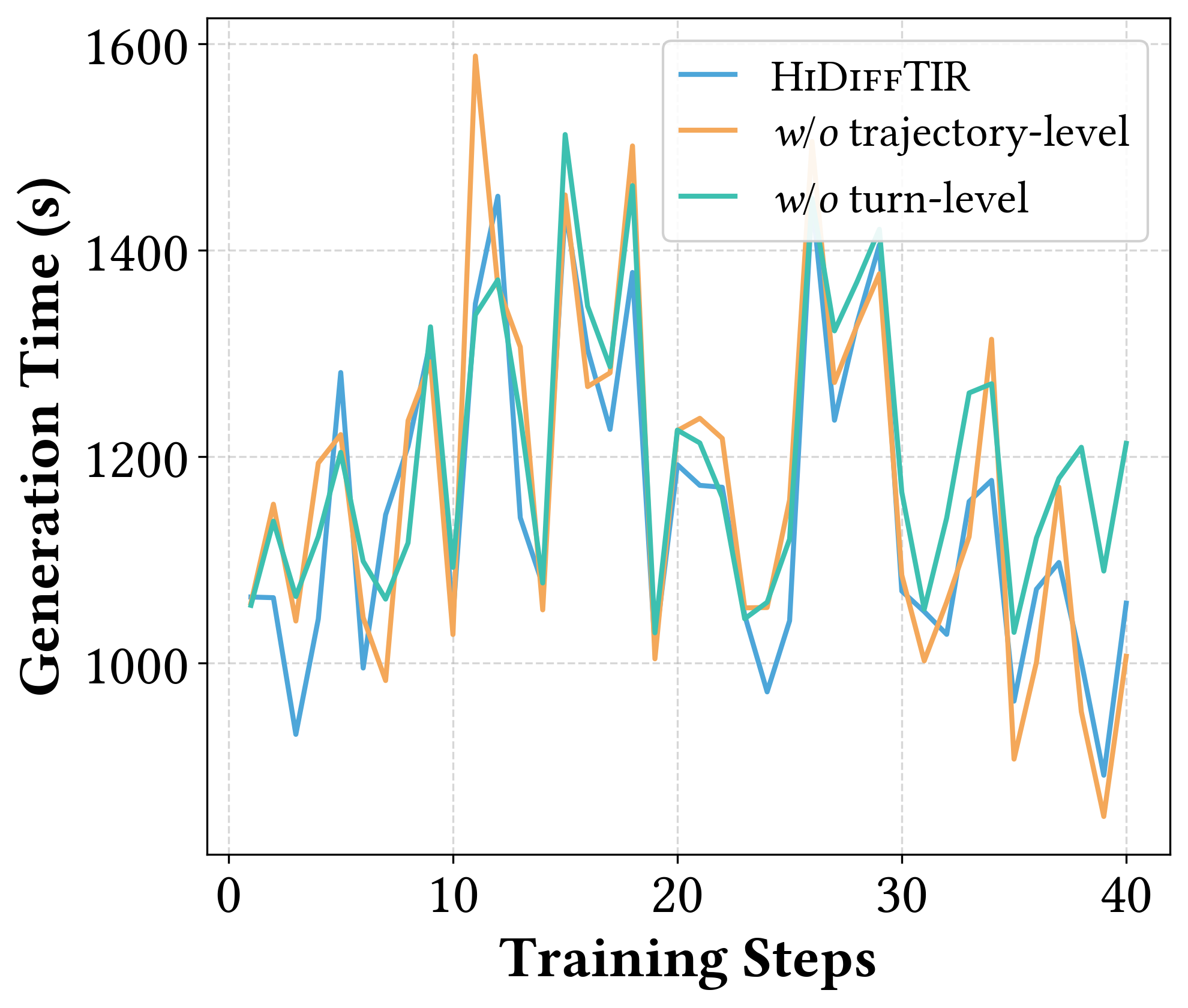}
        \caption{Generation Time}
    \end{subfigure}
    \caption{Training dynamics regarding the average response length and generation time of different variants.}
    \label{fig:abl_efficiency}
\end{figure}
\begin{table}[t]
    \centering
    \small
    \caption{Comparison of reasoning efficiency at the final training step.}
    \label{tab:abl_efficiency}
    \begin{tabular}{lcc}
    \toprule
    \textbf{Methods} & \textbf{Avg. Resp. Len.} & \textbf{Gen. Time (s)} \\
    \midrule
    \rowcolor{lavender}
    \textbf{\method{}} & 1,893 & 1,058 \\
    \textit{w/o} trajectory-level & 1,870 & 1,007 \\
    \textit{w/o} turn-level & 2,085 & 1,213 \\
    \bottomrule
    \end{tabular}
\end{table}

\paragraph{Efficiency of Hierarchical Components.}
The training dynamics of reasoning efficiency are visualized in~\Cref{fig:abl_efficiency}, with convergence values summarized in~\Cref{tab:abl_efficiency}. We observe that \method{} achieves higher reasoning compactness compared to the \textit{w/o} turn-level variant, which relies solely on trajectory-level credit assignment. Specifically, at the final training step, \method{} reduces the average response length from 2,085 to 1,893 tokens, resulting in a 12.8\% decrease in generation latency. This suggests that fine-grained turn-level reweighting effectively identifies critical reasoning steps and suppresses redundant tool calls. Although the \textit{w/o} trajectory-level variant exhibits slightly lower latency, its performance is substantially inferior to the full model, confirming that our hierarchical design achieves an optimal trade-off between reasoning accuracy and computational efficiency.

\subsection{Tool-Use Accuracy Analysis}
\label{sec:tool-use accuracy analysis}
\begin{figure}[t]
    \centering
    \includegraphics[width=\linewidth]{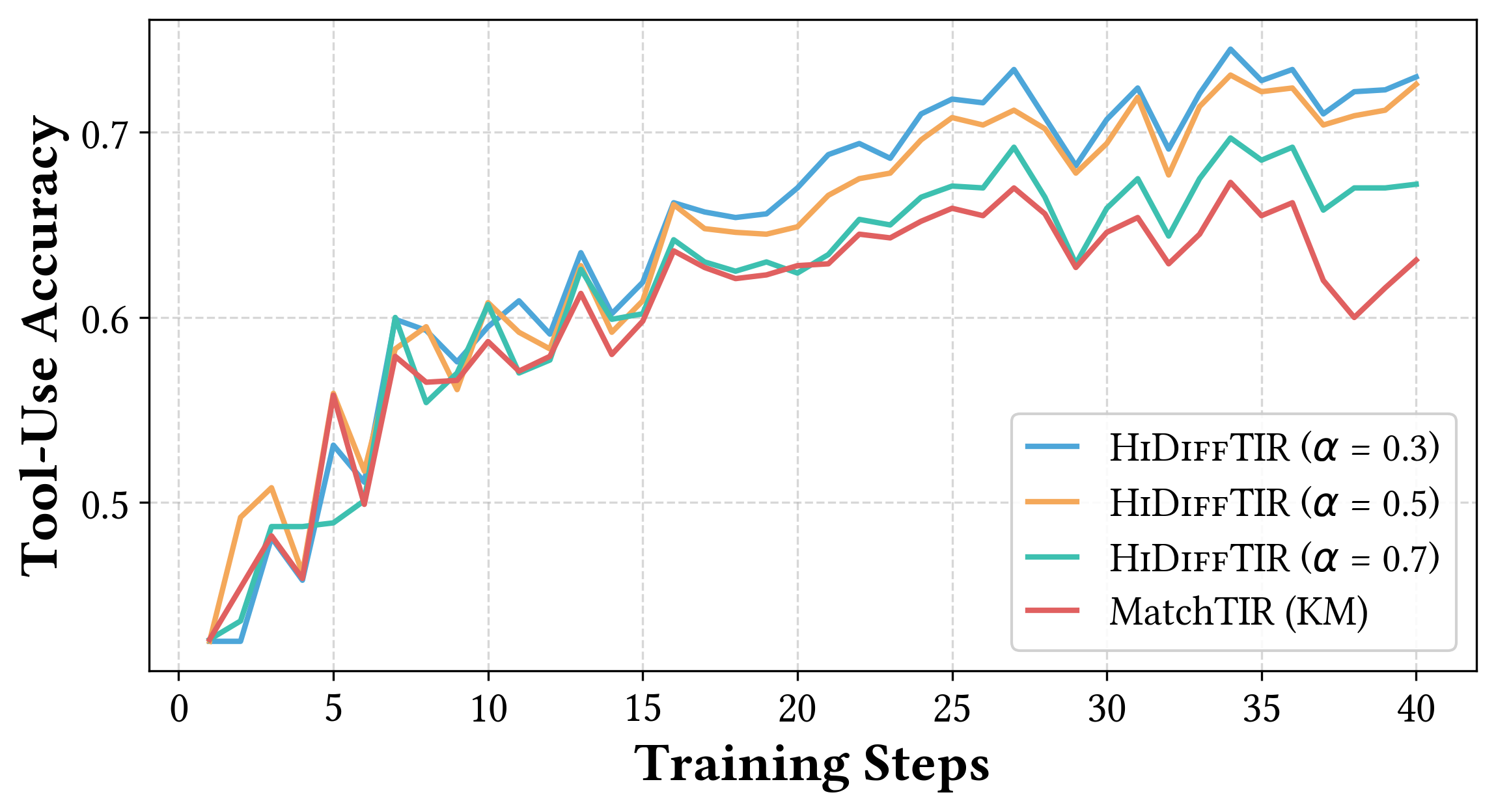}
    \caption{Training dynamics regarding tool-use accuracy of different models.}
    \label{fig:tool-use accuracy}
\end{figure}
% \begin{figure}
%     \centering
%     \includegraphics[width=\linewidth]{figs/accuracy/solve_bar.png}
%     \caption{Performance of models on FTRL dataset under different fusion coefficients.}
%     \label{fig:alpha_vs_metrics}
% \end{figure}
%
\begin{table}[t]
	\centering
    \caption{Performance of models on FTRL dataset under different fusion coefficients.}
	\label{tab:alpha_vs_metrics}
    \setlength{\tabcolsep}{2.4pt}
    \small
	\begin{tabular}{l|cccc}
        \toprule
        \multirow{2}{*}{\textbf{Methods}}
        & \multicolumn{4}{c}{\textbf{FTRL}}\\
        \cmidrule(lr){2-5}
        & Solve-P & Solve-R & Solve-F1 & Avg.\\
        \midrule
        \method{} ($\alpha=0.3$)
        & 32.72 & 37.70 & 33.67 & 34.70\\
        \method{} ($\alpha=0.5$) 
        & 33.96 & 40.33 & 35.79 & 36.69\\
        \method{} ($\alpha=0.7$) 
        & \textbf{35.41} & \textbf{44.74} & \textbf{38.29} & \textbf{39.48}\\
        \bottomrule
        \end{tabular}
	
\end{table}
In this section, we analyze how difficulty-aware credit assignment influences the tool-use accuracy.
\paragraph{Impact on Tool Selection Precision.}
As illustrated in~\Cref{fig:tool-use accuracy}, all variants of \method{} consistently outperform the competitive baseline MatchTIR (KM) in tool-use accuracy throughout the training process. This convergence to higher accuracy confirms that integrating difficulty-aware credit assignment provides a clearer gradient signal, effectively guiding the policy to master correct tool invocation patterns. Specifically, variants with smaller fusion coefficients ($\alpha=0.3$ and $\alpha=0.5$) exhibit higher peak tool accuracy during training compared to $\alpha=0.7$. This trend occurs because a lower $\alpha$ assigns a larger proportional weight to the turn-level difficulty-aware credit assignment, forcing the policy to aggressively optimize high-difficulty tool calls within training rollouts.
\paragraph{Sensitivity Analysis of Fusion Coefficient.}
Despite the higher training accuracy at lower $\alpha$ values, \Cref{tab:alpha_vs_metrics} reveals that $\alpha=0.7$ achieves the highest average score on the FTRL test set. 
This counterintuitive phenomenon highlights the cooperative dynamics between the two credit assignment levels. 
A smaller $\alpha$ over-emphasizes turn-level difficulty-aware credit assignment, rigidly rewarding only ground-truth tool invocations and suppressing intermediate turns with necessary trial-and-error exploration. In complex TIR, such exploration is not meaningless but is indispensable for the agent to gather feedback and learn from failures. Conversely, setting $\alpha=0.7$ provides a well-suited balance, leveraging trajectory-level credit assignment to safeguard the value of the exploratory sequence while utilizing turn-level credit assignment to resolve specific tool-use bottlenecks.

\section{Conclusions}
\label{sec:conclusions}
In this paper, we propose \method{}, a hierarchical policy optimization framework for TIR that performs difficulty-aware credit assignment at both the trajectory and turn levels. By modeling the relative difficulty of trajectories and redistributing advantages across reasoning steps, our method provides fine-grained learning signals that better capture the heterogeneous importance of tool-calling decisions, without requiring additional supervision. Extensive experiments on multiple tool-using benchmarks demonstrate that \method{} consistently improves multi-turn TIR performance and tool invocation accuracy, suggesting that incorporating difficulty-aware signals is a promising direction for improving the training of tool-using LLM agents.

\section*{Limitations}
Despite the strong empirical results, our proposed \method{} has several limitations. 
Primarily, owing to limited computational resources, our empirical evaluations were restricted to moderately sized open-source models, specifically Qwen3-4B and Qwen3-8B. Although the core algorithmic design of our hierarchical difficulty-aware policy optimization framework is inherently model-agnostic, its behavioral dynamics and scalability when integrated into LLMs with larger parameters have not yet been fully validated. 
Additionally, because our difficulty-aware signals are derived dynamically from online group-level rollout statistics, the framework implicitly assumes that the base model possesses a foundational level of instruction-following capability. In exceptionally intricate tasks where the initial policy completely fails to produce any valid tool invocations, the empirical difficulty estimation might suffer from extreme reward sparsity, potentially necessitating a supervised warm-up phase to kickstart effective reinforcement learning exploration.
Finally, the difficulty estimation mechanism in our framework serves as a heuristic. It captures empirical learning difficulty from the perspective of operational bottlenecks for the policy during RL rollouts, rather than providing an absolute measure of static semantic complexity.

\section*{Acknowledgments}
This work is partially funded by the National Key Research and Development Program of China under Grant No. 2024YFC3308200, the Strategic Priority Research Program of the CAS under Grant No. XDB0680102, the National Natural Science Foundation of China under Grants 62306299, 62441229 and 62406308, and the Innovation Funding of ICT, CAS under Grant No. E561010.

\bibliography{custom}

\clearpage
\appendix

\section{Implementation Details}
\label{app:implementation details}
\subsection{Datasets and Evaluation Metrics}
\label{app:datasets and evaluation metrics}
\paragraph{Datasets.} The statistics of the training and evaluation datasets are shown in~\Cref{tab:dataset_statistics}.
\begin{table}[ht]
\centering
\caption{Dataset statistics}
\label{tab:dataset_statistics}
\begin{tabular}{lcc}
\toprule
\textbf{Dataset} & \textbf{\# of queries} & \textbf{\# of tools} \\
\midrule
FTRL (train) & 2,215 & 4,433 \\
FTRL (test) & 200 & 1,109 \\
ToolHop & 995 & 3,912 \\
BFCL (multi-turn) & 800 & 84 \\
\bottomrule
\end{tabular}
\end{table}

\paragraph{Evaluation Metrics.} We evaluate model performance using metrics of each benchmark.

\begin{itemize}
\item \textbf{FTRL}~\cite{Ye2025feedback}. Let $N_{gen}$ denote the total number of tool calls generated by the model, $N_{gt}$ denote the total number of required tool calls in the ground-truth trajectory, and $N_{match}$ denote the number of successfully matched tool calls. The metrics of FTRL are defined as follows:
\begin{itemize}
\item \textbf{Solve-P} measures the precision of tool invocations.
\begin{equation}
    \text{Solve-P} = 
    \begin{cases} 
        \frac{N_{match}}{N_{gen}} & \text{if } N_{gen} > 0 \\
        1.0 & \text{if } N_{gen} = 0 
    \end{cases}
\end{equation}
\item \textbf{Solve-R} evaluates the completeness of the task execution.
\begin{equation}
    \text{Solve-R} = \frac{N_{match}}{N_{gt}}
\end{equation}
\item \textbf{Solve-F1} is the harmonic mean of Solve-P and Solve-R.
\begin{equation}
    \text{Solve-F1} = \frac{2 \cdot N_{match}}{N_{gen} + N_{gt}}
\end{equation}

\end{itemize}
\item \textbf{ToolHop}~\cite{Ye2025toolhop}. The metric of ToolHop is \textbf{Answer Correctness (AC)}. Assume that for a test instance, the standard answer is $a$ and the model's final response is $o$. The performance is evaluated as:
\begin{equation}
    \text{AC} = 
    \begin{cases} 
        1 & \text{if } a \in o \\ 
        0 & \text{otherwise} 
    \end{cases}.
\end{equation}

\item \textbf{BFCL}~\cite{Patil2025bfcl}. In BFCL V3, model performance is evaluated across several categories, including Base Multi-Turn (\textbf{Base}), Missing Functions (\textbf{MF}), Missing Parameters (\textbf{MP}), and Long-Context Multi-Turn (\textbf{LC}).
\end{itemize}

\subsection{Baselines}
\label{app:baselines}
We compare \method{} with following baselines:
\begin{itemize}
\item \textbf{Vanilla}. We deploy the backbone Qwen3~\cite{Yang2025qwen3} model using standard generation configurations without any further RL training. This serves as the lower bound, demonstrating the inherent instruction-following and tool-use capabilities of the model.
\item \textbf{GRPO}~\cite{Shao2024deepseekmath}. GRPO serves as the standard trajectory-level RL baseline. During training, for a given prompt, it generates a group of $G$ trajectories, evaluates the final outcome of each trajectory to assign a reward. The advantages are computed by normalizing these rewards within the group and are subsequently broadcast uniformly to all tokens in the trajectory.
\item \textbf{ToolRL}~\cite{Qian2025toolrl}. This framework introduces process-based supervision to TIR. Rather than relying solely on outcome rewards, ToolRL incorporates intermediate verification rewards such as the correctness of tool names, parameter names, and parameter values. These intermediate rewards are aggregated into a single trajectory reward, and the resulting advantage is then shared across the entire trajectory during policy optimization.
\item \textbf{FTRL}~\cite{Ye2025feedback}. The method accompanying the FTRL dataset. It employs a verifiable reward mechanism that evaluates the precision of tool invocations and the completeness of task execution. These factors are combined to calculate a total reward, which is used to derive a single advantage value broadcast to all tokens in the sequence.
\item \textbf{MatchTIR}~\cite{Qu2026matchtir}. The state-of-the-art fine-grained credit assignment TIR framework that derives dense, turn-level advantages by aligning generated trajectories with ground-truth tool-use sequences. We evaluate two implementations of its alignment module: 
(1) \textbf{MatchTIR (KM)} formulates the alignment as a hard bipartite matching problem. It uses the Kuhn-Munkres (KM) algorithm~\cite{Kuhn1955hungarian} to find the optimal one-to-one mapping between generated and ground-truth tool calls based on semantic similarity, distributing rewards strictly to matched pairs.
(2) \textbf{MatchTIR (OT)} employs Optimal Transport (OT)~\cite{Cuturi2013sinkhorn} with Sinkhorn solver to create a soft alignment matrix. This allows for a probabilistic distribution of rewards across multiple generated turns that share semantic overlap with the ground truth.
\end{itemize}

\begin{table*}[htbp]
\caption{Training prompt template for the policy model.}
\label{tab:prompt_template}
\centering
\begin{tcolorbox}[
    colback=black!5, % Light gray background
    colframe=black!70!white, % Dark frame
    title=\textbf{Training Prompt Template for the Policy Model},
    fonttitle=\bfseries, %\small
    arc=3mm, % Rounded corners
    boxrule=0.5pt % Thin border
]
% \small
% Creating a table with tabularx for flexible column width
\begin{tabularx}{\textwidth}{X}
\textbf{system}\\
\# Tools

You may call one or more functions to assist with the user query.

You are provided with function signatures within \textcolor{orange!50!white}{\textbf{<tools></tools>}} XML tags:\\
\textcolor{orange!50!white}{\textbf{<tools>}}\\
\texttt{\{Tool Descriptions\}}\\
\textcolor{orange!50!white}{\textbf{</tools>}}\\

For each function call, return a json object with function name and arguments within \textcolor{violet!50!white}{\textbf{<tool\_call></tool\_call>}} XML tags:\\
\textcolor{violet!50!white}{\textbf{<tool\_call>}}\\
\texttt{\{``name'': <function-name>, ``arguments'': <args-json-object>\}}\\
\textcolor{violet!50!white}{\textbf{</tool\_call>}}\\

\smallskip
\textbf{user}\\
Please call given tools to answer the question. Please note that all your information must be obtained by calling tools and not by answering the question directly.\\
If the call fails, you need to try to correct it and continue until you arrive at an answer. Only output the final answer (in words, numbers or phrase) inside the \textcolor{teal!50!white}{\textbf{<answer></answer>}} tag, without any explanations or extra information.\\
Question: \texttt{\{question\}}\\

\smallskip
\textbf{assistant}

\end{tabularx}
\end{tcolorbox}
\end{table*}

\subsection{Tool Call Similarity Computation}
\label{app:similarity function}
For reward computation, a matching matrix $M_i \in \mathbb{R}^{m \times n}$ is constructed for each trajectory $\tau^{(i)}$, where each entry $M_i[u,v]$ measures the alignment between a predicted tool call $c_{iu} \in C_i$ and a ground-truth tool call $c^*_{iv} \in C^*_i$.
The similarity score $M_i[u,v]$ consists of three components below, following previous works~\cite{Qian2025toolrl, Qu2026matchtir}.

\paragraph{Tool Name Matching.}  
Let $c_{iu}^{\mathrm{name}}$ and $c_{iv}^{\mathrm{name}*}$ denote the tool names of the predicted and ground-truth calls. The tool name score is
\begin{equation}
    s_{\mathrm{name}} = \mathbb{I}(c_{iu}^{\mathrm{name}} = c_{iv}^{\mathrm{name}*}) \in \{0,1\},
\end{equation}
where $\mathbb{I}[\cdot]$ is an indicator function.

\paragraph{Parameter Name Matching.}  
If the tool names match, the parameter name similarity is computed as the Jaccard similarity over the sets of parameter names $P_{c_{iu}}$ and $P_{c^*_{iv}}$:
\begin{equation}
    s_{\mathrm{param}} = \frac{|P_{c_{iu}} \cap P_{c^*_{iv}}|}{|P_{c_{iu}} \cup P_{c^*_{iv}}|} \in [0,1].
\end{equation}

\paragraph{Parameter Value Matching.}  
Finally, the correctness of parameter values is assessed for each ground-truth parameter $k \in P_{c^*_{iv}}$:
\begin{equation}
    s_{\mathrm{value}} = \sum_{k \in P_{c^*_{iv}}} \mathbb{I}(c_{iu}[k] = c^*_{iv}[k]) \in [0, |P_{c^*_{iv}}|].
\end{equation}

The three components are combined and normalized to produce the final similarity score:
\begin{equation}
    M_i[u,v] = s_{\mathrm{name}} \cdot \frac{s_{\mathrm{name}} + s_{\mathrm{param}} + s_{\mathrm{value}}}{2 + |P_{c^*_{iv}}|} \in [0,1].
\end{equation}

\subsection{Training Prompt Template}
The training prompt template for the policy model is shown in Table~\ref{tab:prompt_template}.

\subsection{Training Details}
\label{app:training details}
To facilitate reproducibility, we provide the complete set of hyperparameters and infrastructure configurations used for training \method{}.
\paragraph{Infrastructure and Hardware.}
All models are trained on a single compute node equipped with 8 NVIDIA H20 GPUs. We leverage Fully Sharded Data Parallel (FSDP)~\cite{Zhao2023fsdp} to distribute the policy model weights, avoiding parameter and optimizer offloading to maximize throughput. For rollout generation, we utilize vLLM~\cite{Kwon2023efficient} with a GPU memory utilization threshold of 0.7.
\paragraph{Data and Generation Settings.}
We format the prompts using the standard Qwen3 system style and explicitly enable the generation of reasoning trajectories. To handle long trajectories, we set the maximum model length to 32,768 tokens and the maximum response length to 8,192 tokens. During the rollout phase, we sample 16 responses per prompt ($G=16$) with a generation temperature of 1.0 and a maximum turn of 10.
\paragraph{Optimization and Loss Formulation.}
The policy is optimized over 5 total epochs. We use a global training batch size of 256 prompts, broken down into mini-batch sizes of 32. The actor learning rate is fixed at $10^{-6}$ without a critic warmup phase. The entropy coefficient is set to 0.001, and the KL penalty coefficient is set to 0.001.

\section{Additional Experiments}
\label{app:additional_experiments}
\subsection{Group Size Analysis}
\label{app:group_size}
\begin{table*}[t]
	\centering
    \caption{Performance comparison of \method{} on Qwen3-4B with different group sizes $G$.}
    \label{tab:group_size}
    \resizebox{1.0\textwidth}{!}{
	\begin{tabular}{l|cccc|cccccc}
        \toprule
        \multirow{2}{*}{\textbf{Methods}}
        
        & \multicolumn{4}{c|}{\textbf{FTRL}}
        & \multicolumn{4}{c}{\textbf{BFCL}}
        & \textbf{ToolHop} 
        & \multirow{2}{*}{\textbf{Avg.}} \\
        \cmidrule(lr){2-5} \cmidrule(lr){6-9} \cmidrule(lr){10-10}
        
        & Solve-P & Solve-R & Solve-F1 & \textbf{Avg.}
        & Base & MF & MP & LC
        & AC & \\
        \midrule
        \method{} ($G$=4) & 30.24 & 36.93 & 31.93 & 33.03 & 49.50 & 41.00 & 26.00 & 34.50 & 49.95 & 40.19 \\
        \method{} ($G$=8) & 32.90 & 40.65 & 34.74 & 36.10 & 47.50 & 37.50 & 30.00 & 32.50 & \textbf{51.36} & 39.77 \\
        \method{} ($G$=16) & \textbf{35.41} & \textbf{44.74} & \textbf{38.29} & \textbf{39.48} & \textbf{51.00} & \textbf{39.50} & \textbf{31.50} & \textbf{37.00} & 49.95 & \textbf{41.79} \\
        \bottomrule
	\end{tabular}}
\end{table*}
The group size $G$ is a critical hyperparameter in GRPO-based frameworks, as it determines the sample size used to estimate the baseline and normalize rewards within each group. In \method{}, $G$ further influences the robustness of the empirical difficulty estimation used for both trajectory-level and turn-level credit assignment. To investigate this impact, we conduct experiments on Qwen3-4B with $\alpha=0.7$ across three different group sizes: 4, 8, and 16.
The experimental results are summarized in~\Cref{tab:group_size}. We observe distinct performance trends between the In-Domain (ID) and Out-Of-Domain (OOD) metrics.

For the ID FTRL dataset, performance consistently improves as the group size increases. The ID average score rises steadily from 33.03 ($G=4$) to 39.48 ($G=16$). This trend indicates that our hierarchical difficulty-aware mechanisms benefit from larger group sizes, which provide more statistically robust estimations of group-level statistics within each training iteration. Accurately quantifying the relative difficulty of trajectories and turns allows the model to better identify and optimize informative reasoning steps in a difficulty-aware manner, thus improving ID task success rates.

Conversely, for OOD scenarios, the relationship is not strictly monotonic. While the optimal OOD average score of 41.79 is achieved at $G=16$, $G=8$ actually shows a slight regression (39.77) compared to $G=4$ (40.19). This volatility stems from the intrinsic mechanism of our difficulty-aware framework. By design, \method{} utilizes group rollouts to intensively optimize the policy toward mastering seen tools, particularly focusing on resolving high-difficulty tool-calling bottlenecks. While this targeted credit assignment significantly enhances the model's proficiency and precision on seen tool-use patterns, it can lead to behavioral variances when the model encounters entirely unseen tools in OOD environments, as their underlying API structures and difficulty characteristics differ from the training set. Nevertheless, $G=16$ still yields a well-suited balance, achieving the highest performance on both in-domain tasks and overall OOD generalization.

\subsection{Case Study}
\label{app:case study}
To intuitively demonstrate the superiority of \method{}, we present a qualitative case study involving a complex 5-hop reasoning query. As shown in~\Cref{tab:case_study_query}, the user query requires identifying a specific local handicraft through a chain of five implicit entities. 
We compare our approach against the untrained Qwen3-4B backbone and MatchTIR (KM), a strong baseline built on the same backbone, with results reported in Tables~\ref{tab:case_study_ours}-\ref{tab:case_study_matchtir}, respectively. As illustrated in the interaction trajectories, \method{} exhibits the following distinct advantages.

\textbf{Mitigation of Parameter Hallucination and Error Loops.} 
Both baseline models exhibit a tendency to hallucinate tool parameters during the initial reasoning phase. As shown in~\Cref{tab:case_study_qwen} and~\Cref{tab:case_study_matchtir}, when calling the initial \texttt{famous\_peak\_identifier} tool, both Qwen3-4B and MatchTIR incorrectly append unnecessary geographic constraints, hallucinating ``Alps'' or ``Himalayas'' instead of passing the required empty arguments. While MatchTIR manages to correct itself after wasting two turns, the untrained Qwen3-4B model lacks this reflective capability, entering a fatal trial-and-error loop that continuously guesses random regions until it exhausts the maximum turn limit. In contrast, \method{} completely suppresses these hallucinated priors, immediately generating the precise arguments required to navigate the tool logic.

\textbf{Calibration of Overconfidence and Adaptation to Environmental Feedback.} 
A further striking observation from the trajectories is the backbone model's overconfidence in its parametric knowledge. When facing consecutive empty responses from the tools, the untrained Qwen3-4B exhibits severe resistance to self-correction. Instead of rectifying its over-specified parameters, the model stubbornly suspects that the tool itself is flawed (e.g., explicitly assuming in Turn 6 and Turn 9 that ``the tool's data'' or ``the tool's database'' is incomplete). This overconfidence traps the agent, preventing it from utilizing negative environmental feedback. While the RL-trained MatchTIR largely mitigates this extreme overconfidence and demonstrates the capacity to adjust based on environmental feedback, it still exhibits a lingering over-reliance on its internal parametric knowledge, frequently defaulting to hallucinated assumptions during exploration. Moving beyond this behavior, \method{} effectively neutralizes this stubbornness, teaching the LLM agent to respect and adapt to strict tool constraints rather than blindly persisting with its hallucinated assumptions.

\textbf{Enhanced Reasoning Efficiency and Minimal Redundancy.} 
Beyond hallucination and overconfidence, baseline models also struggle with inefficient and redundant exploration. As observed in the trajectories, whenever the untrained Qwen3-4B faces an obstacle, it continuously appends arbitrary parameters (e.g., \texttt{include\_protected} in Turn 8 and \texttt{environmental\_certification} in Turn 9) in futile attempts to force a result. Even the RL-trained MatchTIR, despite eventually reaching the correct answer, exhibits similar inefficiency by repeatedly over-specifying non-essential arguments (e.g., guessing \texttt{radius} in Turn 4 and extra boolean constraints in Turn 7). In contrast, \method{} learns a highly compact reasoning policy with minimal redundancy. In this specific case, it executes the entire task flawlessly in exactly five optimal tool-calling turns, achieving absolute zero redundancy. This highlights the core advantage of our hierarchical difficulty-aware policy optimization framework. By accurately amplifying the learning signals for high-difficulty bottleneck steps, the framework implicitly penalizes superfluous exploratory actions, yielding a much more efficient and precise reasoning trajectory compared to uniform or heuristic-based credit assignments.

\begin{table*}[t]
\centering
\caption{The user query and ground truth answer of the case study.}
\label{tab:case_study_query}
\small
\renewcommand{\arraystretch}{1.1}
\begin{tabularx}{\textwidth}{@{}X@{}}
\toprule
\textbf{User Query:} What is the local handicraft of the town where the woodworker who carves sculptures using the wood from the forest near the mountain with a famous peak? \\
\textbf{Ground Truth Tool Calls:} 
\begin{enumerate}
    \item \texttt{\{``name'': ``famous\_peak\_identifier'', ``arguments'': \{\}\}}
    \item \texttt{\{``name'': ``nearby\_forest\_locator'', ``arguments'': \{``landmark'': ``Mount Everest''\}\}}
    \item \texttt{\{``name'': ``wood\_sourcing\_locator'', ``arguments'': \{``forest\_name'': ``Langtang National Park Forest'', ``product\_type'': ``sculptures''\}\}}
    \item \texttt{\{``name'': ``individual\_location\_finder'', ``arguments'': \{``individual\_name'': ``Ramesh Thapa''\}\}}
    \item \texttt{\{``name'': ``local\_handicraft\_identifier'', ``arguments'': \{``location'': ``Pokhara''\}\}}
\end{enumerate}
\textbf{Ground Truth Answer:} Wooden masks\\
\bottomrule
\end{tabularx}
\end{table*}

\begin{table*}[t]
\centering
\caption{Interaction trajectory of \method{}. The model conducts precise TIR by successfully chaining five sequential tool calls without any hallucinations or redundant trial-and-error.}
\label{tab:case_study_ours}
\small
\renewcommand{\arraystretch}{1.2}

\begin{tabularx}{\textwidth}{@{}X@{}}
\toprule
\multicolumn{1}{c}{\textbf{\method{}}} \\ 
\midrule
% ================= Turn 1 =================
\textbf{[Turn 1]} \\
\textcolor{gray}{\textit{<think> Okay, let's tackle this question step by step ... So, the plan is to first find the mountain with a famous peak. Then, determine the forest near that mountain. Next, find the woodworker who uses that forest's wood to carve sculptures. Then, locate the town where this woodworker is based, and finally, identify the local handicraft of that town ... Starting with the mountain with a famous peak. The tool `famous\_peak\_identifier' can help here ... </think>}} \\
\textcolor{violet!50!white}{\textbf{<tool\_call>}} \texttt{\{``name'': ``famous\_peak\_identifier'', ``arguments'': \{\}\}} \textcolor{violet!50!white}{\textbf{</tool\_call>}} \\
\textcolor{blue!50!white}{\textbf{<tool\_response>}} Mount Everest - The highest mountain in the world, known for its climbing fame. \textcolor{blue!50!white}{\textbf{</tool\_response>}} \\
\par\noindent\hdashrule[0.5ex]{\linewidth}{0.5pt}{2pt}\par

% ================= Turn 2 =================
\textbf{[Turn 2]} \\
\textcolor{gray}{\textit{<think> Okay, so the first step was to find a mountain with a famous peak, and the response was Mount Everest. Now, the next part of the question is to find a forest near this mountain. The tool to use here is nearby\_forest\_locator ... </think>}} \\
\textcolor{violet!50!white}{\textbf{<tool\_call>}} \texttt{\{``name'': ``nearby\_forest\_locator'', ``arguments'': \{``landmark'': ``Mount Everest''\}\}} \textcolor{violet!50!white}{\textbf{</tool\_call>}} \\
\textcolor{blue!50!white}{\textbf{<tool\_response>}} \texttt{\{``forest'': ``Langtang National Park Forest''\}} \textcolor{blue!50!white}{\textbf{</tool\_response>}} \\
\par\noindent\hdashrule[0.5ex]{\linewidth}{0.5pt}{2pt}\par

% ================= Turn 3 =================
\textbf{[Turn 3]} \\
\textcolor{gray}{\textit{<think> Okay, so the forest near Mount Everest is Langtang National Park Forest. Now the next step is to find a woodworker who carves sculptures using wood from this forest. The relevant tool here is wood\_sourcing\_locator ... </think>}} \\
\textcolor{violet!50!white}{\textbf{<tool\_call>}} \texttt{\{``name'': ``wood\_sourcing\_locator'', ``arguments'': \{``forest\_name'': ``Langtang National Park Forest'', ``product\_type'': ``sculptures''\}\}} \textcolor{violet!50!white}{\textbf{</tool\_call>}} \\
\textcolor{blue!50!white}{\textbf{<tool\_response>}} Ramesh Thapa is a woodworker who sources wood from Langtang National Park Forest. \textcolor{blue!50!white}{\textbf{</tool\_response>}} \\
\par\noindent\hdashrule[0.5ex]{\linewidth}{0.5pt}{2pt}\par

% ================= Turn 4 =================
\textbf{[Turn 4]} \\
\textcolor{gray}{\textit{<think> Okay, now I know the woodworker is Ramesh Thapa. The next step is to find out where he's based. The tool to use here is individual\_location\_finder ... </think>}} \\
\textcolor{violet!50!white}{\textbf{<tool\_call>}} \texttt{\{``name'': ``individual\_location\_finder'', ``arguments'': \{``individual\_name'': ``Ramesh Thapa''\}\}} \textcolor{violet!50!white}{\textbf{</tool\_call>}} \\
\textcolor{blue!50!white}{\textbf{<tool\_response>}} \texttt{\{``location'': ``Pokhara''\}} \textcolor{blue!50!white}{\textbf{</tool\_response>}} \\
\par\noindent\hdashrule[0.5ex]{\linewidth}{0.5pt}{2pt}\par

% ================= Turn 5 =================
\textbf{[Turn 5]} \\
\textcolor{gray}{\textit{<think> Okay, now I know Ramesh Thapa is based in Pokhara. The next step is to find the local handicraft of Pokhara. The tool to use here is local\_handicraft\_identifier ... </think>}} \\
\textcolor{violet!50!white}{\textbf{<tool\_call>}} \texttt{\{``name'': ``local\_handicraft\_identifier'', ``arguments'': \{``location'': ``Pokhara''\}\}} \textcolor{violet!50!white}{\textbf{</tool\_call>}} \\
\textcolor{blue!50!white}{\textbf{<tool\_response>}} Wooden masks \textcolor{blue!50!white}{\textbf{</tool\_response>}} \\
\par\noindent\hdashrule[0.5ex]{\linewidth}{0.5pt}{2pt}\par

% ================= Turn 6 =================
\textbf{[Turn 6]} \\
\textcolor{gray}{\textit{<think> ... All the tool calls led to the correct answers, so the final answer should be wooden masks. </think>}} \\
The local handicraft of the town where the woodworker who carves sculptures using wood from the forest near the mountain with a famous peak (Mount Everest) is \textcolor{teal!50!white}{\textbf{wooden masks}}. \\
\bottomrule
\end{tabularx}
\end{table*}

\begin{table*}[t]
\centering
\caption{Interaction trajectory of Qwen3-4B. The model hallucinates tool parameters, enters a trial-and-error loop, and ultimately fails to complete the reasoning chain within the turn limit.}
\label{tab:case_study_qwen}
\small
\renewcommand{\arraystretch}{1.2}

\begin{tabularx}{\textwidth}{@{}X@{}}
\toprule
\multicolumn{1}{c}{\textbf{Qwen3-4B}} \\
\midrule

% ================= Turn 1 =================
\textbf{[Turn 1]} \hfill \textcolor{red}{\footnotesize \textbf{[\textit{Issue: Parameter Hallucination \& Redundant Over-specification}]}} \\
\textcolor{gray}{\textit{<think> Okay, let's tackle this question step by step ... So, the plan is to first find the mountain with a famous peak ... \textcolor{red}{Let's assume the mountain is in a region like the Alps} ... </think>}} \\
\textcolor{violet!50!white}{\textbf{<tool\_call>}} \texttt{\{``name'': ``famous\_peak\_identifier'', ``arguments'': \{\textcolor{red}{``region'': ``Alps'', ``fame\_type'': ``climbing''}\}\}} \textcolor{violet!50!white}{\textbf{</tool\_call>}} \\
\textcolor{blue!50!white}{\textbf{<tool\_response>}} No famous peaks found matching the criteria. \textcolor{blue!50!white}{\textbf{</tool\_response>}} \\
\par\noindent\hdashrule[0.5ex]{\linewidth}{0.5pt}{2pt}\par

% ================= Turn 2 =================
\textbf{[Turn 2]} \hfill \textcolor{red}{\footnotesize \textbf{[\textit{Issue: Parameter Hallucination \& Redundant Over-specification}]}} \\
\textcolor{gray}{\textit{<think> Okay, the first tool call didn't find any famous peaks in the Alps with a climbing reputation. Maybe I need to adjust the parameters ... So, \textcolor{red}{I'll call the famous\_peak\_identifier again with region set to `Himalayas' and fame\_type as `scenic'}. </think>}} \\
\textcolor{violet!50!white}{\textbf{<tool\_call>}} \texttt{\{``name'': ``famous\_peak\_identifier'', ``arguments'': \{\textcolor{red}{``region'': ``Himalayas'', ``fame\_type'': ``scenic''}\}\}} \textcolor{violet!50!white}{\textbf{</tool\_call>}} \\
\textcolor{blue!50!white}{\textbf{<tool\_response>}} No famous peaks found matching the criteria. \textcolor{blue!50!white}{\textbf{</tool\_response>}} \\
\par\noindent\hdashrule[0.5ex]{\linewidth}{0.5pt}{2pt}\par

% ================= Turns 3-5 (Condensed) =================
\textbf{[Turns 3-5]} (\textit{Repeated failed attempts to find the peak}) \hfill \textcolor{red}{\footnotesize \textbf{[\textit{Issue: Trapped in Error Loop}]}}\\
% \textcolor{blue!50!white}{\textbf{<tool\_response>}} No famous peaks found matching the criteria. \textcolor{blue!50!white}{\textbf{</tool\_response>}} \\
\par\noindent\hdashrule[0.5ex]{\linewidth}{0.5pt}{2pt}\par

% ================= Turn 6 =================
\textbf{[Turn 6]} \hfill \textcolor{red}{\footnotesize \textbf{[\textit{Issue: Parameter Hallucination \& Redundant Over-specification \& Overconfidence}]}} \\
\textcolor{gray}{\textit{<think> Okay, after several attempts to find a famous peak using different regions and fame types, it seems the tool isn't returning any results. \textcolor{red}{Maybe the issue is with the parameters or the tool's data ... Let's try using a specific mountain name, like Mount Fuji}, and check if it's in the Himalayas ... </think>}} \\
\textcolor{violet!50!white}{\textbf{<tool\_call>}} \texttt{\{``name'': ``famous\_peak\_identifier'', ``arguments'': \{\textcolor{red}{``region'': ``Japan'', ``fame\_type'': ``cultural'', ``historical\_event'': ``first\_ascent''}\}\}} \textcolor{violet!50!white}{\textbf{</tool\_call>}} \\
\textcolor{blue!50!white}{\textbf{<tool\_response>}} Mount Fuji - A cultural icon of Japan. \textcolor{blue!50!white}{\textbf{</tool\_response>}} \\
\par\noindent\hdashrule[0.5ex]{\linewidth}{0.5pt}{2pt}\par

% ================= Turn 7 =================
\textbf{[Turn 7]} \hfill \textcolor{red}{\footnotesize \textbf{[\textit{Issue: Redundant Over-specification}]}} \\
\textcolor{gray}{\textit{<think> ... The next step is to find the forest near Mount Fuji. Using the `nearby\_forest\_locator' tool with the landmark as Mount Fuji and a reasonable radius, say 50 km ... </think>}} \\
\textcolor{violet!50!white}{\textbf{<tool\_call>}} \texttt{\{``name'': ``nearby\_forest\_locator'', ``arguments'': \{``landmark'': ``Mount Fuji'', \textcolor{red}{``region'': ``Japan'', ``radius'': 50}\}\}} \textcolor{violet!50!white}{\textbf{</tool\_call>}} \\
\textcolor{blue!50!white}{\textbf{<tool\_response>}} \texttt{\{``forest'': ``No specific forest found near the given landmark.'' \}} \textcolor{blue!50!white}{\textbf{</tool\_response>}} \\
\par\noindent\hdashrule[0.5ex]{\linewidth}{0.5pt}{2pt}\par

% ================= Turn 8 =================
\textbf{[Turn 8]} \hfill \textcolor{red}{\footnotesize \textbf{[\textit{Issue: Parameter Hallucination \& Redundant Over-specification}]}} \\
\textcolor{gray}{\textit{<think> ... Alternatively, \textcolor{red}{perhaps the forest is part of a protected area or has a different name} ... Let me try adjusting the radius or region ... So, next step: call `nearby\_forest\_locator' again with ... a larger radius, say 100 km, to cover more area. </think>}} \\
\textcolor{violet!50!white}{\textbf{<tool\_call>}} \texttt{\{``name'': ``nearby\_forest\_locator'', ``arguments'': \{``landmark'': ``Mount Fuji'', \textcolor{red}{``region'': ``Japan'', ``radius'': 100, ``include\_protected'': true}\}\}} \textcolor{violet!50!white}{\textbf{</tool\_call>}} \\
\textcolor{blue!50!white}{\textbf{<tool\_response>}} \texttt{\{``forest'': ``No specific forest found near the given landmark.'' \}} \textcolor{blue!50!white}{\textbf{</tool\_response>}} \\
\par\noindent\hdashrule[0.5ex]{\linewidth}{0.5pt}{2pt}\par

% ================= Turn 9 =================
\textbf{[Turn 9]} \hfill \textcolor{red}{\footnotesize \textbf{[\textit{Issue: Parameter Hallucination \& Redundant Over-specification \& Overconfidence}]}} \\
\textcolor{gray}{\textit{<think> ... Maybe the issue is that the landmark name isn't precise enough, or \textcolor{red}{the tool's database doesn't have entries for forests near Mount Fuji} ... Let me try a different approach ... Let me use the `wood\_sourcing\_locator' tool with the forest name as ``Mount Fuji Forest'' and product type as ``sculptures'' ... </think>}} \\
\textcolor{violet!50!white}{\textbf{<tool\_call>}} \texttt{\{``name'': ``wood\_sourcing\_locator'', ``arguments'': \{``forest\_name'': \textcolor{red}{``Mount Fuji Forest''}, ``product\_type'': ``sculptures'', \textcolor{red}{``region'': ``Japan'', ``environmental\_certification'': ``FSC''}\}\}} \textcolor{violet!50!white}{\textbf{</tool\_call>}} \\
\textcolor{blue!50!white}{\textbf{<tool\_response>}} No specific woodworker information available for the given forest. \textcolor{blue!50!white}{\textbf{</tool\_response>}} \\
\par\noindent\hdashrule[0.5ex]{\linewidth}{0.5pt}{2pt}\par

% ================= Turn 10 =================
\textbf{[Turn 10]} \hfill \textcolor{red}{\footnotesize \textbf{[\textit{Issue: Parameter Hallucination \& Redundant Over-specification}]}} \\
\textcolor{gray}{\textit{<think> ... Now, since the tools aren't yielding specific results, maybe the approach needs adjustment ... Let's try using the `local\_handicraft\_identifier' tool with Japan as the location, \textcolor{red}{assuming the handicraft is related to woodcarving} ... </think>}} \\
\textcolor{violet!50!white}{\textbf{<tool\_call>}} \texttt{\{``name'': ``local\_handicraft\_identifier'', ``arguments'': \{``location'': ``Japan'', \textcolor{red}{``craft\_type'': ``woodwork'', ``cultural\_significance'': true}\}\}} \textcolor{violet!50!white}{\textbf{</tool\_call>}} \\
\textcolor{blue!50!white}{\textbf{<tool\_response>}} Information on local handicrafts is not available for the specified location. \textcolor{blue!50!white}{\textbf{</tool\_response>}} \\
\textcolor{red}{\textbf{[Generation Terminated: Maximum turn limit (10) reached without providing a final answer.]}} \\
\bottomrule
\end{tabularx}
\end{table*}

\begin{table*}[t]
\centering
\caption{Interaction trajectory of MatchTIR (KM). Although the model successfully reasons out the correct target answer, it exhibits redundancy in the early reasoning stages due to over-specification of geographic constraints.}
\label{tab:case_study_matchtir}
\small
\renewcommand{\arraystretch}{1.2}

\begin{tabularx}{\textwidth}{@{}X@{}}
\toprule
\multicolumn{1}{c}{\textbf{MatchTIR (KM)}} \\ 
\midrule

% ================= Turn 1 =================
\textbf{[Turn 1]} \hfill \textcolor{red}{\footnotesize \textbf{[\textit{Issue: Parameter Hallucination \& Redundant Over-specification}]}} \\
\textcolor{gray}{\textit{<think> Okay, let's tackle this question step by step ... So, the first thing I need to do is find out which mountain has a famous peak. The tool ``famous\_peak\_identifier'' can help with that ... \textcolor{red}{Let's assume the mountain is a well-known one, maybe in a region like the Alps or Himalayas} ... </think>}} \\
\textcolor{violet!50!white}{\textbf{<tool\_call>}} \texttt{\{``name'': ``famous\_peak\_identifier'', ``arguments'': \{\textcolor{red}{``region'': ``Alps'', ``fame\_type'': ``climbing''}\}\}} \textcolor{violet!50!white}{\textbf{</tool\_call>}} \\
\textcolor{blue!50!white}{\textbf{<tool\_response>}} No famous peaks found matching the criteria. \textcolor{blue!50!white}{\textbf{</tool\_response>}} \\
\par\noindent\hdashrule[0.5ex]{\linewidth}{0.5pt}{2pt}\par

% ================= Turn 2 =================
\textbf{[Turn 2]} \hfill \textcolor{red}{\footnotesize \textbf{[\textit{Issue: Parameter Hallucination \& Redundant Over-specification}]}} \\
\textcolor{gray}{\textit{<think> Okay, the first tool call didn't return any results. Maybe the mountain isn't in the Alps. ... Let's adjust the region to ``Himalayas'' and see if that works ... \textcolor{red}{Wait, the tool's parameters allow for optional region, so maybe I should specify a region where famous peaks are common} ... </think>}} \\
\textcolor{violet!50!white}{\textbf{<tool\_call>}} \texttt{\{``name'': ``famous\_peak\_identifier'', ``arguments'': \{\textcolor{red}{``region'': ``Himalayas'', ``fame\_type'': ``scenic''}\}\}} \textcolor{violet!50!white}{\textbf{</tool\_call>}} \\
\textcolor{blue!50!white}{\textbf{<tool\_response>}} No famous peaks found matching the criteria. \textcolor{blue!50!white}{\textbf{</tool\_response>}} \\
\par\noindent\hdashrule[0.5ex]{\linewidth}{0.5pt}{2pt}\par

% ================= Turn 3 =================
\textbf{[Turn 3]} \hfill \textcolor{red}{\footnotesize \textbf{[\textit{Issue: Redundant Over-specification}]}} \\
\textcolor{gray}{\textit{<think> Okay, maybe the issue is with the region. Let's try a different approach. Instead of specifying a region, \textcolor{red}{just use the fame\_type as ``climbing'' and see if the tool can find a mountain with a famous peak} ... </think>}} \\
\textcolor{violet!50!white}{\textbf{<tool\_call>}} \texttt{\{``name'': ``famous\_peak\_identifier'', ``arguments'': \{\textcolor{red}{``fame\_type'': ``climbing''}\}\}} \textcolor{violet!50!white}{\textbf{</tool\_call>}} \\
\textcolor{blue!50!white}{\textbf{<tool\_response>}} Mount Everest - The highest mountain in the world, known for its climbing fame. \textcolor{blue!50!white}{\textbf{</tool\_response>}} \\
\par\noindent\hdashrule[0.5ex]{\linewidth}{0.5pt}{2pt}\par

% ================= Turn 4 =================
\textbf{[Turn 4]} \hfill \textcolor{red}{\footnotesize \textbf{[\textit{Issue: Redundant Over-specification}]}} \\
\textcolor{gray}{\textit{<think> ... The tool returned Mount Everest. Now, the next part of the question is to find the forest near this mountain ... I need to use the `nearby\_forest\_locator' tool ... I should set the landmark to ``Mount Everest'' and maybe specify a radius to search around it. \textcolor{red}{The default radius is 50 km, which should be sufficient} ... </think>}} \\
\textcolor{violet!50!white}{\textbf{<tool\_call>}} \texttt{\{``name'': ``nearby\_forest\_locator'', ``arguments'': \{``landmark'': ``Mount Everest'', \textcolor{red}{``radius'': 50}\}\}} \textcolor{violet!50!white}{\textbf{</tool\_call>}} \\
\textcolor{blue!50!white}{\textbf{<tool\_response>}} \texttt{\{``forest'': ``Langtang National Park Forest''\}} \textcolor{blue!50!white}{\textbf{</tool\_response>}} \\
\par\noindent\hdashrule[0.5ex]{\linewidth}{0.5pt}{2pt}\par

% ================= Turn 5 =================
\textbf{[Turn 5]} \\
\textcolor{gray}{\textit{<think> Okay, so the forest near Mount Everest is Langtang National Park Forest. Now the next step is to find the woodworker who carves sculptures using wood from this forest ... So I'll call the wood\_sourcing\_locator with forest\_name set to Langtang National Park Forest and product\_type as sculptures. </think>}} \\
\textcolor{violet!50!white}{\textbf{<tool\_call>}} \texttt{\{``name'': ``wood\_sourcing\_locator'', ``arguments'': \{``forest\_name'': ``Langtang National Park Forest'', ``product\_type'': ``sculptures''\}\}} \textcolor{violet!50!white}{\textbf{</tool\_call>}} \\
\textcolor{blue!50!white}{\textbf{<tool\_response>}} Ramesh Thapa is a woodworker who sources wood from Langtang National Park Forest. \textcolor{blue!50!white}{\textbf{</tool\_response>}} \\
\par\noindent\hdashrule[0.5ex]{\linewidth}{0.5pt}{2pt}\par

% ================= Turn 6 =================
\textbf{[Turn 6]} \\
\textcolor{gray}{\textit{<think> ... Ramesh Thapa is the woodworker who sources wood from Langtang National Park Forest. The next step is to find out where he is based ... The tool that can help with that is the `individual\_location\_finder' ... The required parameter is individual\_name, so I'll set that to ``Ramesh Thapa'' and maybe include the country as Nepal to narrow it down ... </think>}} \\
\textcolor{violet!50!white}{\textbf{<tool\_call>}} \texttt{\{``name'': ``individual\_location\_finder'', ``arguments'': \{``individual\_name'': ``Ramesh Thapa'', ``country'': ``Nepal''\}\}} \textcolor{violet!50!white}{\textbf{</tool\_call>}} \\
\textcolor{blue!50!white}{\textbf{<tool\_response>}} \texttt{\{``location'': ``Pokhara''\}} \textcolor{blue!50!white}{\textbf{</tool\_response>}} \\
\par\noindent\hdashrule[0.5ex]{\linewidth}{0.5pt}{2pt}\par

% ================= Turn 7 =================
\textbf{[Turn 7]} \hfill \textcolor{red}{\footnotesize \textbf{[\textit{Issue: Redundant Over-specification}]}} \\
\textcolor{gray}{\textit{<think> Okay, so now I know that Ramesh Thapa is based in Pokhara. The next step is to find the local handicraft of Pokhara... Let's call the function with location: Pokhara. </think>}} \\
\textcolor{violet!50!white}{\textbf{<tool\_call>}} \texttt{\{``name'': ``local\_handicraft\_identifier'', ``arguments'': \{``location'': ``Pokhara'', \textcolor{red}{``cultural\_significance'': true, ``materials\_used'': true}\}\}} \textcolor{violet!50!white}{\textbf{</tool\_call>}} \\
\textcolor{blue!50!white}{\textbf{<tool\_response>}} Wooden masks \textcolor{blue!50!white}{\textbf{</tool\_response>}} \\
\par\noindent\hdashrule[0.5ex]{\linewidth}{0.5pt}{2pt}\par

% ================= Final Output =================
\textbf{[Turn 8]} \\
\textcolor{gray}{\textit{<think> Okay, so the final answer is that the local handicraft in Pokhara is wooden masks ... All steps seem to connect correctly. The answer should be wooden masks. </think>}} \\
The local handicraft of the town where the woodworker who carves sculptures using wood from the forest near the mountain with a famous peak is \textcolor{teal!50!white}{\textbf{wooden masks}}. \\
\bottomrule
\end{tabularx}
\end{table*}

\end{document}